\documentclass[sigconf,screen=true,anonymous=false,bookmarks=false]{acmart}

\usepackage{enumitem}
\setlist{leftmargin=5.5mm}

\iftrue
\fi

\usepackage{lipsum}
\usepackage{graphicx}
\usepackage{amsmath}
\usepackage{footnote}
\usepackage{mathtools}
\usepackage{mathrsfs}     
\usepackage{comment}
\usepackage[subrefformat=parens,labelformat=parens]{subfig}
\usepackage{bm,bbm}
\usepackage{multirow}
\usepackage{threeparttable,booktabs}
\usepackage{blkarray}
\usepackage{tikz}
\usetikzlibrary{positioning,calc,fit,decorations.pathmorphing,shapes.geometric,shapes.gates.logic.US,calc}
\usepackage{balance}
\usepackage{courier}                                       
\usepackage{makecell}
\usepackage{cleveref}                                      

\usepackage[mathcal]{eucal}
\usepackage[]{algpseudocode}                               
\algrenewcommand\textproc{\texttt}
\makeatletter\let\float@addtolists\relax\makeatother
\usepackage{algorithm}
\usepackage{filecontents}                                  
\usepackage{pgfplots}
\usepackage{pgfplotstable}
\usepackage{pgf-pie}
\usepgfplotslibrary{groupplots}
\pgfplotsset{compat=newest}
\usepackage[figuresright]{rotating}
\usepackage{xcolor,colortbl}                               

\theoremstyle{plain}

\theoremstyle{definition}

\algrenewcommand\textproc{\texttt}

\usepackage[skip=1pt]{caption}            
\definecolor{CUHKorange}{RGB}{244,106,18} 
\definecolor{CUHKblue}{RGB}{0,111,190}    
\definecolor{CUHKgreen}{RGB}{0,127,128}   
\definecolor{CUHKred}{RGB}{228,46,36}     
\definecolor{CUHKyellow}{RGB}{198,148,34} 
\definecolor{CUHKdark}{RGB}{114,44,114}   
\definecolor{CUHKmiddle}{RGB}{144,44,144} 

\usepackage{tcolorbox}
\tcbuselibrary{skins,breakable}
    {\endtcolorbox}

\usepackage{graphicx}
\graphicspath{{pdf/}}
\usepackage{enumitem}
\copyrightyear{2026}
\acmYear{2026}
\setcopyright{cc}
\setcctype{by}
\acmConference[DAC '26]{63rd ACM/IEEE Design Automation Conference}{July 26--29, 2026}{Long Beach, CA, USA}
\acmBooktitle{63rd ACM/IEEE Design Automation Conference (DAC '26), July 26--29, 2026, Long Beach, CA, USA}
\acmDOI{10.1145/3770743.3804259}
\acmISBN{979-8-4007-2254-7/2026/07}

\renewcommand{\shortauthors}{}                       
\newcommand{\shorttitle}{}                         
\fancypagestyle{standardpagestyle}{
  \fancyhf{}

}
\fancypagestyle{firstpagestyle}{
  \fancyhf{}

}
\makeatletter
\let\@copyrightspace\relax
\let\@teaserfig\relax
\let\@teaserblank\relax
\makeatother
\begin{document}
\title{OpenACMv2: An Accuracy-Constrained Co-Optimization Framework for Approximate DCiM}

\thanks{This work was supported by the Fundamental Research Funds for the Central Universities (Grant Nos. 30925010605 and 30924012004), the National Key Laboratory of Integrated Circuits and Microsystems (Grant No. JCYQ2310803-1), and the Jiangsu Provincial Major Science and Technology Project (Grant No. BG2025012). $^*$Corresponding authors. }

\author{Yiqi Zhou$^1$,Yue Yuan$^1$,Yikai Wang$^1$,Bohao Liu$^2$,Qinxin Mei$^2$,Zhuohua Liu$^3$,Shan Shen$^{1,*}$,Wei Xing$^{4,*}$,Daying Sun$^{1,*}$,Li Li$^1$, and Guozhu Liu$^5$}

\affiliation{
  \institution{$^1$Nanjing University of Science and Technology, China, $^2$Shenzhen University, China}
  \institution{$^3$Beihang University, China, $^4$University of Sheffield, UK}
  \institution{$^5$The 58th Research Institute of China Electronics Technology Group Corporation, China}
  \country{}
}

\email{{shanshen, hasdysun}@njust.edu.cn, w.xing@sheffield.ac.uk}

\begin{abstract}
Digital Compute-in-Memory (DCiM) accelerates neural networks by reducing data movement. \textit{Approximate} DCiM can further improve power–performance–area (PPA), but demands accuracy-constrained co-optimization across coupled architecture and transistor-level choices. Building on OpenYield, we introduce \textit{Accuracy-Constrained Co-Optimization (ACCO)} and present \textbf{OpenACMv2}, an open framework that operationalizes ACCO via \textit{two-level} optimization: (1) accuracy-constrained architecture search of compressor combinations and SRAM macro parameters, driven by a fast GNN-based surrogate for PPA and error; and (2) variation- and PVT-aware transistor sizing for standard cells and SRAM bitcells using Monte Carlo. By decoupling ACCO into architecture-level exploration and circuit-level sizing, OpenACMv2 integrates classic single- and multi-objective optimizers to deliver strong PPA–accuracy tradeoffs and robust convergence. The workflow is compatible with FreePDK45 and OpenROAD, supporting reproducible evaluation and easy adoption. Experiments show that the proposed two-level ACCO framework achieves most of its accuracy-constrained efficiency gain at Level-1 through architecture exploration, delivering roughly 50\%+ PDP reduction, while Level-2 transistor-level optimization provides a further single-digit PDP improvement while preserving accuracy, enabling rapid “what-if” exploration for approximate DCiM. The framework is available on \href{https://github.com/ShenShan123/OpenACM}{GitHub}.
\vspace{-6pt}
\end{abstract}

\maketitle
\thispagestyle{empty}
\pagestyle{empty}

\section{Introduction}

Digital Compute-in-Memory (DCiM) accelerates neural network (NN) workloads by performing low-precision arithmetic in/near SRAM, reducing data movement and interconnect switching. This collapses the von Neumann bottleneck and delivers strong power–performance–area (PPA) gains while remaining compatible with standard digital design flows~\cite{hardware_approximate_techniques}.

Approximate computing further amplifies these gains: many NNs tolerate moderate error without accuracy loss~\cite{approxann_framework,exploiting_approximate_computing}. Leveraging approximate multipliers and compressors can significantly reduce power and area, but introduces a hard co-design problem—balancing accuracy (e.g., Mean Relative Error Distance (MRED)/Normalized Mean Error Distance (NMED)) and PPA across coupled architecture and transistor choices. These trade-offs are nonconvex, PVT/variation-sensitive, and strongly influenced by SRAM configuration, making manual tuning impractical.
\begin{figure}[tbp]
\vspace{4pt}
	\centering
	\includegraphics[width=\linewidth]{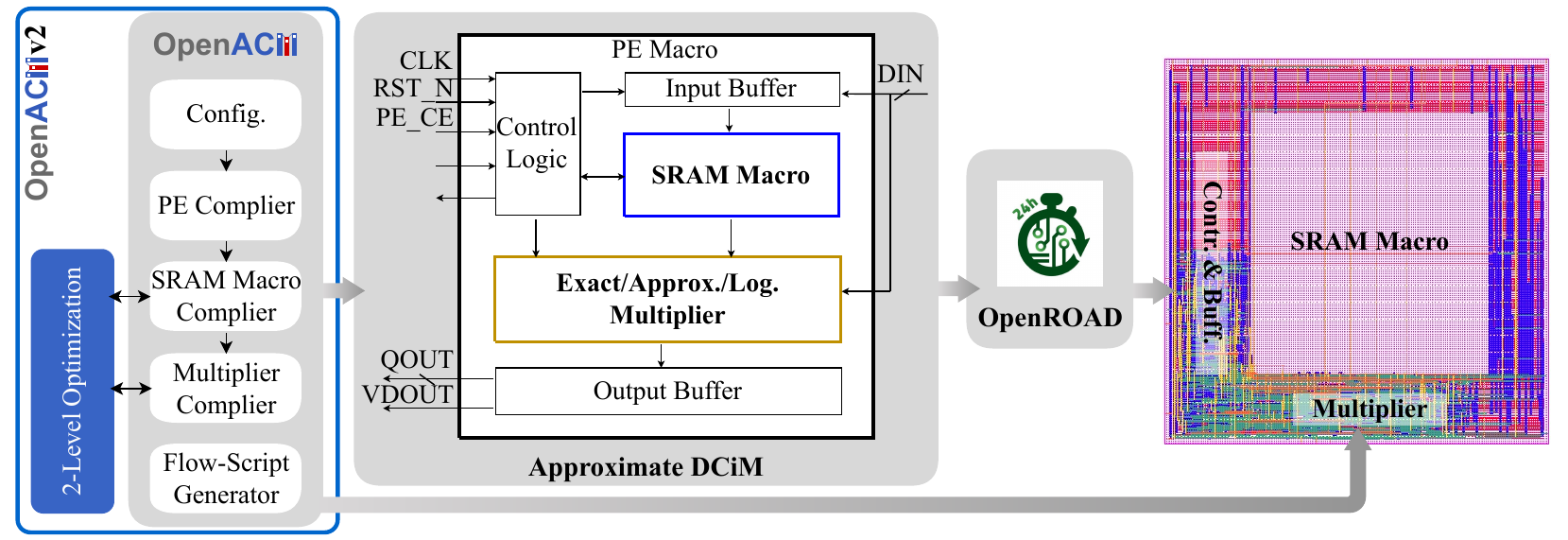}
	\caption{The overall structure of OpenACM, an approximate DCiM compiler, provides an end-to-end DCiM flow. A processing element contains a multiplier and an SRAM macro.}
    \label{fig:acm_intro}
\end{figure}
An automatic compiler is essential. OpenACM~\cite{openacm}, the first open-source SRAM-based approximate CiM compiler, provides an end-to-end DCiM flow (see \autoref{fig:acm_intro}), but lacks a unified framework that jointly optimizes architecture exploration and transistor sizing under explicit accuracy with PVT/variation awareness. More broadly, AutoDCIM~\cite{autodcim2023}, SynDCIM~\cite{syndcim2024}, ARCTIC~\cite{arctic2024}, and OpenC\textsuperscript{2}~\cite{openc2_2025} target exact arithmetic and remain application-agnostic, without accuracy-aware modeling or rapid evaluation of approximate arithmetic. As a result, ACCO-style loops—which require millions of ``what-if'' queries—are infeasible; even a single 16-bit query can take tens of seconds with conventional synthesis/simulation. Compounding the challenge, approximate DCiM exposes more knobs than exact flows: compressor topology and bit-slicing, partial-product truncation and gating, SRAM bank configuration, and transistor sizing across both standard cells and SRAM cells. These choices interact through variation and PVT. Today, there is still no unified, open optimization framework that jointly reasons about accuracy, topology, and sizing, while providing fast evaluation of approximate multipliers within an ACCO loop; prior DCiM-specific optimization studies do not address this joint architecture–transistor co-optimization under accuracy constraints.

To bridge this gap, we introduce \textit{Accuracy-Constrained Co-Optimization (ACCO)}, which elevates application-level accuracy budgets to first-class constraints in the front-end design loop. Building on OpenYield~\cite{openyield} and OpenACM~\cite{openacm}, \textbf{OpenACMv2} operationalizes ACCO with a two-level flow: architecture exploration under accuracy budgets followed by variation/PVT-aware circuit sizing for both standard cells and memory cells. OpenACMv2 integrates a GNN surrogate of approximate multiplier performance for fast design space exploration, and supports classic single- and multi-objective optimizers. This \textit{top-down divide-and-conquer} strategy enables efficient search, strong PPA–accuracy tradeoffs, and robust convergence. The workflow is fully open-source and compatible with FreePDK45~\cite{freepdk45} and OpenROAD~\cite{openroad}.
OpenACMv2 makes the following key contributions:
\begin{itemize}
    \item \textbf{Two-Level Optimization for DCiM:} We co-optimize \textit{architecture-level} selections (compressor combinations, SRAM bank configuration) under explicit application accuracy constraints, and \textit{circuit-level} transistor sizing for both standard and memory cells, running Monte Carlo simulations across PVT variations.

    \item \textbf{Algorithmic Integration:} We incorporate a suite of single-objective and multi-objective optimizers and demonstrate that classic algorithms achieve strong results when ACCO is decoupled into architecture-level and circuit-level sub-stages for both logic and memory cell customizations.

    \item \textbf{Fast GNN-Based Approximate Multiplier Modeling:} We construct \textit{PEA-GNN}, a performance model to rapidly evaluate diverse approximate multiplier designs, providing orders-of-magnitude speedup for PPA and error queries while preserving fidelity needed for accuracy-aware decisions.
    
\end{itemize}


\section{Background}
\label{sec:background}
Neural networks exhibit strong tolerance to numerical noise, where moderate perturbations typically do not affect task-level accuracy~\cite{approxann_framework,exploiting_approximate_computing,hardware_approximate_techniques}. 
This property enables trading arithmetic precision for improved PPA efficiency when computation is executed near or inside SRAM arrays. 
Within DCiM systems, such approximation yields substantial energy benefits while remaining compatible with standard digital design flows~\cite{sram_cim_review}.

Despite these advantages, designing a DCiM processing engine (PE) that meets explicit accuracy budgets while optimizing PPA remains highly challenging. 
Multipliers dominate both computational error and energy in DCiM PEs. 
Decades of work propose diverse approximate 4--2 compressors and multipliers with varied error/PPA profiles~\cite{momeni2014design,yang2015approximate,akbari2017dual,ha2017multipliers,strollo2020comparison,kong2021design}; Kim and Del~Barrio~\cite{KIM2024104364} show that combining complementary compressors can reduce bias, but only in restricted settings.

DCiM introduces two architectural degrees of freedom that significantly expand the design space and enable fine-grained trade-offs: 
(1) assigning multiple types of approximate compressors across partial-product (PP) columns, and 
(2) selecting which PP columns use approximation versus exact compressors (e.g., approximating lower-significance columns while keeping higher-significance ones exact, see \autoref{fig:mul_structure}). 
This flexibility allows constructing multipliers with tunable accuracy and energy to meet application-level constraints. 
However, exploring this large combinatorial space lacks fast, high-fidelity evaluation. 
Accurate PPA/error estimation typically requires logic synthesis, timing analysis, and power simulation per configuration, which prevents efficient inner-loop search.

Beyond architectural choices, approximate multipliers can be further improved via transistor-level optimization. 
For each compressor family, device sizing, gate decomposition, and threshold-voltage assignment (LVT/HVT) can reduce delay, energy, and area. 
In practice, this leads to customized standard-cell implementations rather than relying solely on off-the-shelf gates; architecture–circuit co-optimization then pushes energy efficiency toward the Pareto front.

On the memory side, SRAM macro parameters—row/column splits, array counts, and mux ratios—govern access delay and energy under a fixed capacity budget~\cite{sram_cim_review}. 
Bitcell sizing further tunes performance but also affects PVT sensitivity and yield stability. 
Combined architecture–transistor co-optimization across macro parameters and sizing achieves the best PPA for the DCiM macro.

Together, heterogeneous compressor assignment, selective PP-column approximation, transistor sizing, and SRAM configuration create a large, nonconvex design space that existing optimization algorithms cannot handle at scale. 
Meanwhile, most frameworks focus on either transistor sizing or architecture-level optimization and lack accuracy-aware evaluation for DCiM systems. In practice, optimizing architecture and transistor parameters concurrently yields similar PPA gains to the two-level, top–down ACCO strategy, but significantly increases complexity and slows convergence. 
We therefore adopt a two-level approach that decouples architecture exploration from device sizing under explicit accuracy budgets.

These challenges motivate treating the DCiM PE as a first-class optimization target. 
OpenACMv2 implements ACCO with a top–down divide-and-conquer strategy: 
(1) architecture exploration over compressor combinations, PP-column approximation, and SRAM macro parameters,  
(2) variation/PVT-aware transistor sizing for logic and memory cells, and  
(3) fast performance prediction via a PEA-GNN surrogate to avoid expensive simulations.

\section{Framework of \raisebox{-3pt}{\includegraphics[height=13pt]{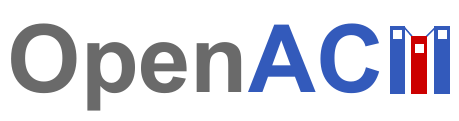}}v2}
\label{sec:overview}

\begin{figure}[tbp]
	\centering
	\includegraphics[width=\linewidth]{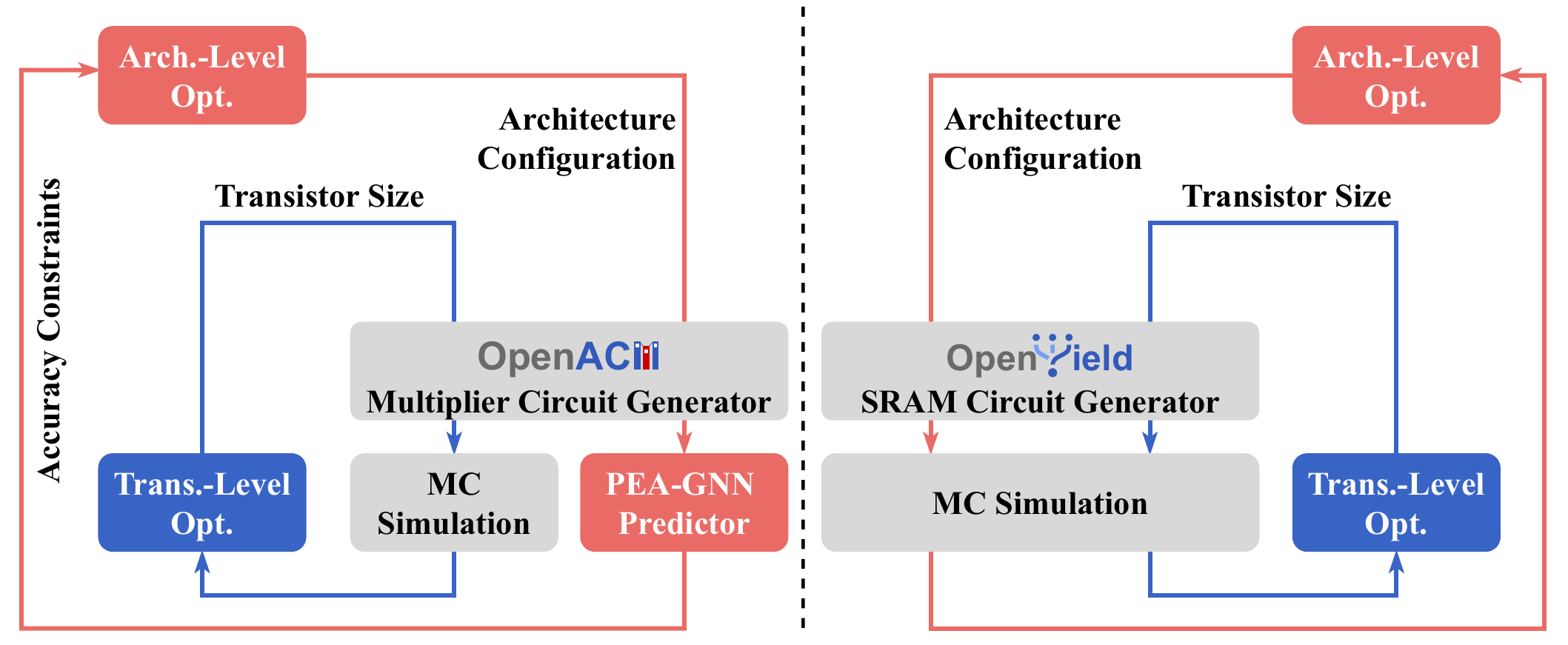}
	\caption{OpenACMv2 ACCO workflow: Level-1 architecture search guided by PEA-GNN, followed by Level-2 variation/PVT-aware transistor sizing. Foundation circuit generators are OpenACM and OpenYield.}
    \label{fig:framework}
\end{figure}

\autoref{fig:framework} shows a two-level, top–down workflow across the approximate multiplier and SRAM macro. The \textit{top level} rapidly explores architectures using lightweight models and a GNN surrogate to screen candidates. The \textit{bottom level} performs variation/PVT-aware transistor tuning to refine shortlisted designs. 

\textbf{Approximate Multiplier Optimization.}
At the top level, we identify compressor assignments and column-level approximation patterns using efficient search guided by a graph neural network surrogate; we rapidly estimate PPA and error to rank candidates.
At the bottom level, we perform variation/PVT-aware transistor sizing for the selected compressors to preserve functional correctness while further optimizing PDP and area and enhancing robustness across multiple corners.

\textbf{SRAM Macro Optimization.}
The top level determines macro configurations under capacity budgets using SPICE simulation to obtain PPA and prune dominated candidates.
Bottom level applies variation/PVT-aware sizing for bitcells to ensure stability and timing.
This two-level, divide-and-conquer flow couples efficient search with a GNN surrogate, scales to large design spaces, and yields energy-efficient, variation-aware DCiM under accuracy constraints.

\section{Conquering the Approximate Multiplier}

Having outlined the two-level flow, we now focus on the approximate multiplier. We begin with the architecture-level configuration space and later refine selected compressors at the transistor level.

\subsection{Top-Level: Approximate Compressor Combinations}

\begin{figure}[tbp]
	\centering
	\includegraphics[width=\linewidth]{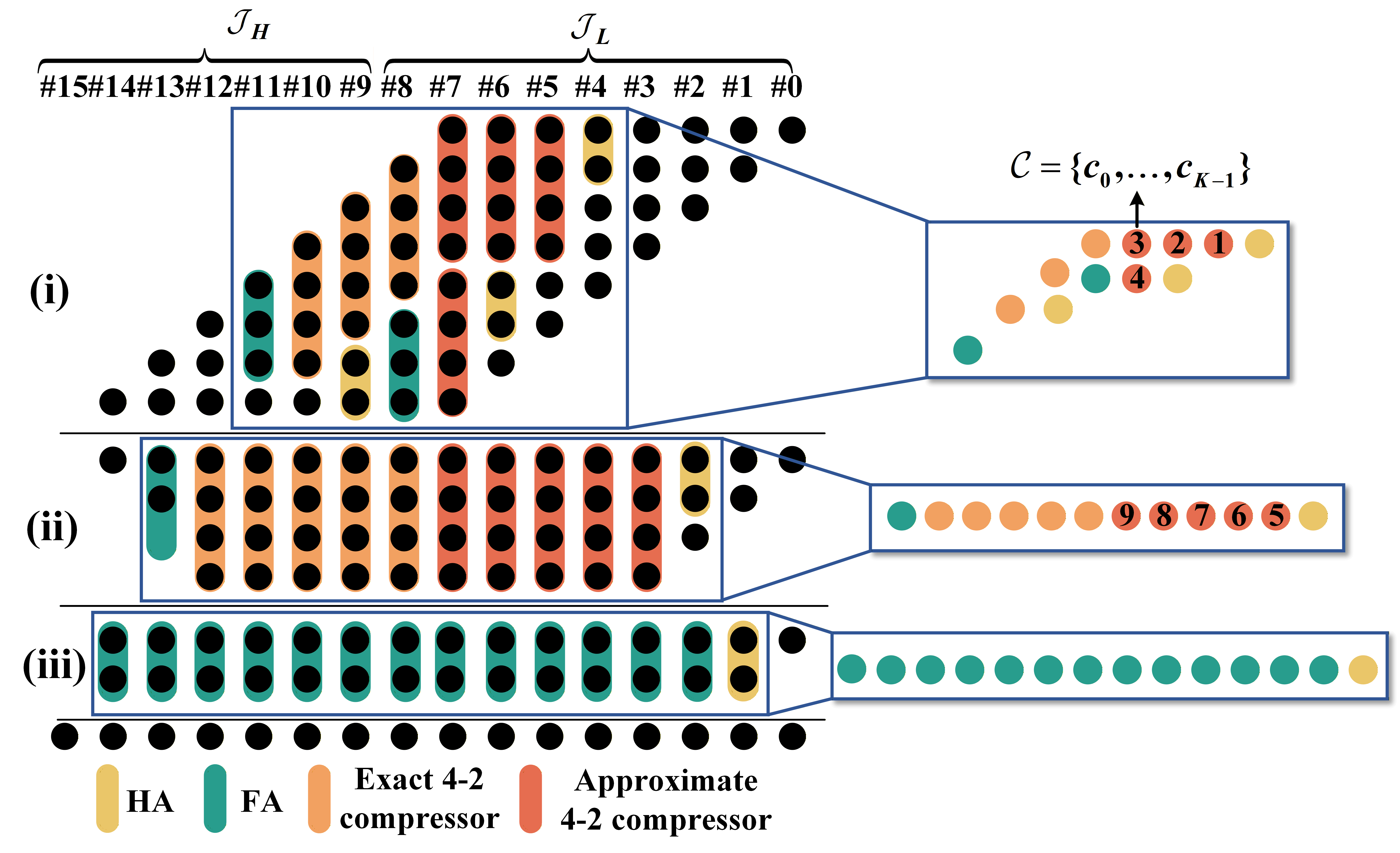}
	\caption{The structure of the 8-bit approximate multiplier. It consists of 3 stages: (i) partial-product generation, (ii) a configurable reduction tree with approximate compressors on selected columns, and (iii) a final carry-propagate adder.}
    \label{fig:mul_structure}
\end{figure}

Building on \autoref{fig:mul_structure}, we now formalize the architecture-level configuration space.

\label{sec:app}

\subsubsection{Design Parameters}

For the $N$-bit multiplier in Fig.~\ref{fig:mul_structure}, partial products (PPs) form
$2N$ columns indexed by $\mathcal{J}=\{0,\ldots,2N-1\}$. Approximation is restricted
to the lower significant columns $\mathcal{J}_L=\{0,\ldots,N-1\}$, while the remaining
columns $\mathcal{J}_H=\{N,\ldots,2N-1\}$ remain exact.

Within $\mathcal{J}_L$, the layout provides $t=9$ configurable compressor positions, corresponding to the indexed positions shown in Fig.~\ref{fig:mul_structure}. Each slot $i\in\{1,\ldots,t\}$
lies at column $j_i\in\mathcal{J}_L$, and may choose one compressor from the
library $\mathcal{C}=\{c_0,\ldots,c_{K-1}\}$ with $K=8$ types. A design is thus
encoded by $\mathbf{a}=[a_1,\ldots,a_t]$, where $a_i\in\{0,\ldots,K-1\}$ denotes
the selected compressor type. The architecture-level search space therefore
contains $K^t = 8^9$ possible combinations.

This parameterization explicitly captures:
(i) the selective approximation region defined by $\mathcal{J}_L$, and
(ii) heterogeneous compressor assignment through the vector $\mathbf{a}$.

\subsubsection{Objective Functions}

Next, we define the accuracy and physical metrics used for evaluation and the top-level objective.
Given a configuration $\mathbf{a}$, we define the exact multiplication result $R_{\text{exact}}(x,y)$ and the approximate result $R_{\mathbf{a}}(x,y)$ for inputs $(x,y) \in \{0,\ldots,2^N{-}1\}^2$. Let the error distance be $\operatorname{ED}_{\mathbf{a}}(x,y){=}|R_{\mathbf{a}}(x,y){-}R_{\text{exact}}(x,y)|$.

MRED is computed over the subset $\mathcal{U}_+{=}\{(x,y):R_{\text{exact}}(x,y){>}0\}$ to avoid division by zero:
\begin{equation}
\text{MRED}(\mathbf{a}) 
\;=\; \frac{1}{|\mathcal{U}_+|} \sum_{(x,y)\in\mathcal{U}_+} \frac{\operatorname{ED}_{\mathbf{a}}(x,y)}{R_{\text{exact}}(x,y)}.
\end{equation}
Here, $\operatorname{ED}_{\mathbf{a}}$ is the absolute arithmetic error, and MRED measures average \emph{relative} error magnitude; $\text{MRED}{=}0$ indicates exact computation.
NMED normalizes by the maximum result $R_{\max}{=}(2^N{-}1)^2$:
\begin{equation}
\text{NMED}(\mathbf{a}) 
\;=\; \frac{1}{|\mathcal{U}|} \sum_{(x,y)\in\mathcal{U}} \frac{\operatorname{ED}_{\mathbf{a}}(x,y)}{R_{\max}},
\end{equation}
where $\mathcal{U}$ is the input set used for evaluation (uniform grid or application distribution); $\text{NMED}\in[0,1]$ provides an absolute error scale. Physical metrics are delay $D(\mathbf{a})$, dynamic power $P(\mathbf{a})$, and area $A(\mathbf{a})$, with power--delay product $\operatorname{PDP}(\mathbf{a}){=}P(\mathbf{a})\cdot D(\mathbf{a})$. Here $D$ is worst-case path delay (50\% threshold), $P$ captures capacitive switching under the workload toggle rates, and $A$ is cell area.

With these metrics, the top-level optimizer solves the following multi-objective problem:
\begin{equation}
\min_{\mathbf{a}\in\{0,\ldots,K{-}1\}^t} \; \big(\, \text{MRED}(\mathbf{a}),\; \operatorname{PDP}(\mathbf{a}) \,\big)
\quad \text{s.t.}\; \text{NMED}(\mathbf{a}) \leq \varepsilon_{\text{NMED}},
\end{equation}
with optional accuracy constraints (MRED or NMED) depending on the application.

\subsubsection{PEA-GNN: Performance Model of Multiplier}

To evaluate configurations efficiently, PEA-GNN encodes a compressor assignment $\mathbf{a}$ into a stage-wise graph $G{=}(V,E)$ aligned with the multiplier’s compression hierarchy. Nodes are combinational logic circuits (HA, FA, exact/approximate 4--2 compressors); edges are signal interconnects. Stages $\{S_1,\ldots,S_T\}$ partition $V$ by reduction level ($T$ equals the number of compression stages).

\textbf{Graph/node features.} For a node with $n$ inputs, $m$ outputs and variant $y$, let the truth table be $\mathbf{H}^{(y)}\in\{0,1\}^{2^n\times m}$ (digital mapping of inputs to outputs). With independent inputs of marginals $\{p_j\}_{j=1}^n$ (bit-1 probabilities; typically $p_j{=}0.5$ under uniform toggling), define the input-combination weights
\begin{equation}
\omega_i 
\;=\; \prod_{j=1}^{n} p_j^{b_{i,j}} (1{-}p_j)^{1-b_{i,j}}, \quad b_i\in\{0,1\}^n,
\end{equation}
and the input-probability vector of each nodes $\bm{p}_{\text{in}} = (\mathbf{H}^{(y)})^\top \bm{\omega}$. Given an error vector $\bm{e}^{(y)}$ (bit-level deviation from the exact compressor), the fused node feature is
\begin{equation}
\bm{h}_v^{(k)} \;=\; \bm{p}_{\text{in}} \odot \bm{e}^{(y)}.
\end{equation}

\textbf{Stage-wise message passing.} Within each stage subgraph $G[S_k]$, GraphSAGE updates embeddings by
\begin{equation}
\bm{h}_v^{(k{+}1)} \;=\; \sigma\!\Big( \mathbf{W}_k\big[\, \bm{h}_v^{(k)} \| \operatorname{AGG}_k(\{\bm{h}_u^{(k)}: u\in \mathcal{N}(v)\}) \big] \Big),
\end{equation}
where $\sigma$ is a nonlinearity, $\mathbf{W}_k$ are learnable weights, and $\operatorname{AGG}_k$ is a neighborhood aggregator (e.g., mean). This propagates localized information along the compression flow. After $T$ stages, critical nodes $\mathcal{V_\text{critical}}{=}V_{\text{approx}}\cup V_{\text{last}}$ are concatenated to form a global representation $\bm{g}$.

\textbf{Regression head.} A shared MLP maps $\bm{g}$ to predictions
\begin{equation}
\widehat{\bm{y}} \;=\; \big( \widehat{\text{MRED}},\widehat{\text{NMED}},\widehat{D},\widehat{A},\widehat{P} \big),
\end{equation}
with Softplus on outputs to ensure non-negativity (consistent with physical metrics).
\begin{figure}[tbp]
    \centering
    \includegraphics[width=\linewidth]{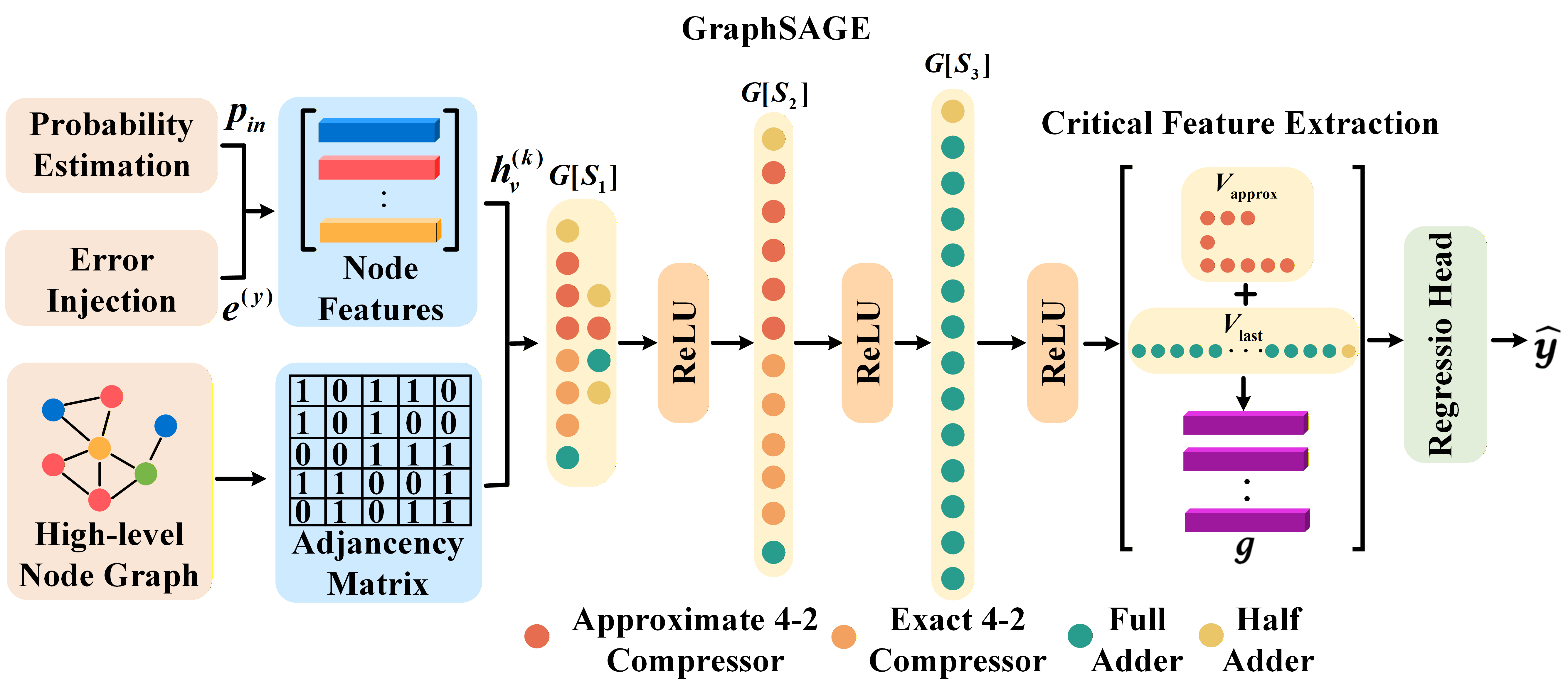}
    \caption{PEA-GNN: stage-wise graph, node-feature fusion, and regression.}
    \label{GNN}
\end{figure}

As shown in \autoref{GNN}, this hierarchical design yields fast, high-fidelity predictions for accuracy and PPA while aligning message passing with the compression stages.

\subsubsection{Multi-Objective Optimization Algorithms}
Equipped with the surrogate, we adopt the same iterative view as above. At iteration $t$, select the next feasible compressor configuration $\mathbf{x}_{t{+}1}$ either by scoring candidates with $a_t(\mathbf{x})$ or by applying an update operator $U_t$; stochastic methods may accept a proposal with probability $A_t$. Let the objectives be $\mathbf{f}(\mathbf{x}){=}\big(\text{MRED}(\mathbf{x}),\operatorname{PDP}(\mathbf{x})\big)$. We say $\mathbf{x}\prec\mathbf{y}$ if $f_i(\mathbf{x})\le f_i(\mathbf{y})$ for all $i$ and strictly for some $j$; the Pareto set is $\mathcal{X}^*$ and the frontier $\mathcal{F}^*$.
(i)~\textbf{MOEA/D}~\cite{zhang2007moea}: scalarization-based selection $\mathbf{x}_{t{+}1}{=}\arg\min_{\mathbf{x}} s_{\boldsymbol{\lambda}}(\mathbf{x})$ (weighted Tchebycheff or weighted sum); subproblems cooperate to approximate $\mathcal{F}^*$.
(ii)~\textbf{NSGA-II}~\cite{deb2002fast}: population update $\mathbf{x}_i^{t{+}1}{=}U_t(\mathbf{x}_i^t)$ with non-dominated sorting and crowding-distance selection to preserve diversity along $\mathcal{F}^*$.
(iii)~\textbf{SMAC}~\cite{hutter2011sequential}: surrogate-guided acquisition $\mathbf{x}_{t{+}1}{=}\arg\max_{\mathbf{x}} a_t\big(s_{\boldsymbol{\lambda}}(\mathbf{x})\big)$ under random-forest models; effective in categorical spaces.
(iv)~\textbf{MOBO}~\cite{gelbart2014bayesian,yang2019multi}: acquisition-based selection maximizing expected hypervolume improvement; balances exploration and exploitation toward higher-Pareto-quality solutions.

Together with PEA-GNN, these backends rapidly generate accuracy--PDP Pareto frontiers without costly Electronic Design Automation (EDA), improving architecture-level exploration throughput. We now transition to bottom-level transistor tuning for compressor implementations.

\subsection{Bottom-Level: Standard Cell Customization}
\label{sec:comp_transistor}

Finally, complementing the top-level exploration, transistor-level optimization is performed for each approximate 4--2 compressor based on its CMOS gate-level implementation (technology standard cells: \texttt{INV}, \texttt{NAND/NOR}, \texttt{XOR/XNOR}, \texttt{AOI/OAI}, \texttt{AND/OR}, \texttt{BUF}, etc.).

\subsubsection{Design Parameters}
We first define the design variables and grouping used for SPICE evaluation. Channel length $L$ is fixed; transistor \emph{widths} are the design variables. For compressor type $y\in\{1,\ldots,8\}$, let $\mathcal{G}_y{=}\{g_1,\ldots,g_{q_y}\}$ be width-sharing groups of equivalent devices (e.g., pull-up/pull-down pairs). The design vector is $\mathbf{w}{=}(w_1,\ldots,w_{q_y})\in\mathbb{R}_+^{q_y}$ with bounds $w_k\in[w_k^{\min},w_k^{\max}]$. Grouping preserves symmetry and reduces dimensionality; the instantiated netlist $\mathcal{N}_y(\mathbf{w})$ is evaluated by circuit-level SPICE simulation.

\subsubsection{Objective Functions}
We now state the correctness condition and physical metrics, then the bottom-level objective. Let $\mathbf{H}^{(y)}\in\{0,1\}^{2^n\times m}$ be the target truth table of compressor $y$, where $n$ and $m$ are the numbers of inputs and outputs, respectively. For input $b_i\in\{0,1\}^n$, let $\mathbf{V}_{\text{out}}(\mathbf{w};b_i)\in\mathbb{R}^m$ be simulated output voltages of $\mathcal{N}_y(\mathbf{w})$, and $\operatorname{c}(\cdot)$ the logic classification using technology thresholds $(V_{\text{OL}}^{\max},V_{\text{OH}}^{\min})$. Functional correctness requires
\begin{equation}
\operatorname{c}\big(\mathbf{V}_{\text{out}}(\mathbf{w};b_i)\big) \;=\; \mathbf{H}^{(y)}_{i,:}\quad \forall\, i\in\{1,\ldots,2^n\}.
\end{equation}
We define delay $D(\mathbf{w})$ as worst-case path delay (50\% threshold), dynamic power $P(\mathbf{w})$ from transient toggling, and area $A(\mathbf{w})$ from cell-area models; $\operatorname{PDP}(\mathbf{w}){=}P(\mathbf{w})\cdot D(\mathbf{w})$. Putting these together, the bottom-level sizing problem is:
\begin{equation}
\begin{aligned}
\min_{\mathbf{w}\in\mathcal{W}_y}\quad  \big( \operatorname{PDP}(\mathbf{w}),&\, A(\mathbf{w}) \big) \\
\text{s.t.}\quad \operatorname{c}\big(\mathbf{V}_{\text{out}}(\mathbf{w}; b_i)\big) = \mathbf{H}^{(y)}_{i,:}, &\ \forall i, w_k \in [w_k^{\min}, w_k^{\max}]\,.
\end{aligned}
\end{equation}
Optionally, robustness can be enforced across PVT corners $\mathcal{C}$ by replacing metrics with their worst-case values, e.g., $D^{\max}(\mathbf{w}){=}\max_{c\in\mathcal{C}}D(\mathbf{w};c)$.

This transistor-level refinement complements Level~I architecture by co-optimizing device sizes with compressor assignment, improving energy efficiency and implementation robustness of approximate multipliers.

\section{Conquering SRAM Macro}

Having refined multiplier logic, we now focus on the SRAM macro. We begin with bank-level configuration under a capacity constraint and later tune bitcells at the transistor level.

\subsection{Top-Level: SRAM Bank Configuration}
\label{sec:sram_architecture}

In parallel with multiplier exploration, we use the open-source \textbf{OpenYield} framework for variation-aware analysis. Here we formalize the SRAM architecture-level configuration space and notation.

\subsubsection{Design Parameters}
We first define the discrete array organization under a capacity constraint. OpenYield explores the architectural design space of SRAM macros subject to fixed storage capacity.
The objective is to identify an array organization—defined by $(r, c, \mu, n_{\!a})$ with rows $r$, columns $c$, mux ratio $\mu$, and number of arrays $n_{\!a}$—that balances dynamic performance and bank area.
All feasible configurations satisfy the capacity constraint:
\begin{equation}
r \times c \times n_{\!a} = \text{Capacity}
\end{equation}
where \textit{Capacity} denotes total storage bits. In this section, we set $\text{Capacity}{=}4\,\text{KB}\times 8{=}32{,}768$ bits, and derive $n_{\!a}$ to ensure full utilization. Candidates follow power-of-two grids: $r\in\{2,4,8,\ldots,512\}$ and $c\in\{2,4,8,\ldots,256\}$, while $\mu$ is selected from a technology-defined set.

\subsubsection{Objective Function}

Next, given a valid configuration $(r,c,\mu,n_{\!a})$, we evaluate dynamic performance and bank area using nominal transistor parameters. Let $D_{\text{rd}}$ and $D_{\text{wr}}$ be read/write access delays, and $P_{\text{rd}}$ and $P_{\text{wr}}$ be read/write powers. Define
\begin{equation}
\begin{aligned}
D_{\max} &= \max\{D_{\text{rd}},\,D_{\text{wr}}\},\quad
P_{\max} = \max\{P_{\text{rd}},\,P_{\text{wr}}\}, \\
A_{\text{bank}}&(r,c,\mu,n_{\!a}) = \text{estimated area of the bank for }(r,c,\mu, n_{\!a}).
\end{aligned}
\end{equation}
To unify the optimization target across both hierarchical levels, we define the figure of merit (FOM) that emphasizes low power, small area, and short access delay:
\begin{equation}
\text{FOM}(r,c,\mu,n_{\!a}) \;=\; -\,\log_{10}\!\left(P_{\max}\,\sqrt{A_{\text{bank}}}\,D_{\max}\right)
\end{equation}
Here, larger FOM indicates better dynamic efficiency for the given organization. With these metrics, architecture-level maximizes $\text{FOM}(r,c,\mu)$ over the feasible set subject to the capacity constraint above and the discrete candidate grids. Scanning candidates yields power–delay Pareto curves that visualize energy–speed trade-offs across array organizations and bank configurations by FOM. This architectural exploration sets up the algorithmic comparison below.

\subsubsection{Single-Objective Optimization Algorithms}
Equipped with the FOM, we use a unified iterative view: at iteration $t$, select the next feasible configuration $\mathbf{x}_{t{+}1}$ either by scoring candidates with a function $a_t(\mathbf{x})$ or by applying an update operator $U_t$ to the current state; stochastic methods may accept a proposal with probability $A_t$. We minimize $f(\mathbf{x}){=}{-}\,\text{FOM}(\mathbf{x})$ subject to capacity and technology constraints.
(i)~\textbf{CBO}~\cite{cbo2014}: Acquisition-based selection $\mathbf{x}_{t{+}1}{=}\arg\max_\mathbf{x}\, a_t(\mathbf{x})\,\pi_t(\mathbf{x})$, where $a_t$ encodes improvement under a GP model and $\pi_t$ promotes feasibility; uncertainty balances exploration and exploitation.
(ii)~\textbf{PSO}~\cite{pso1995}: Population update $\mathbf{x}_i^{t{+}1}{=}U_t(\mathbf{x}_i^t, \mathcal{B}_t)$, with inertia and attraction toward personal/global bests $\mathcal{B}_t$; particles coordinate toward high-FOM regions.
(iii)~\textbf{SA}~\cite{sa1983}: Neighborhood proposal $\mathbf{x}'\sim\mathcal{N}(\mathbf{x}_t)$ with acceptance $A_t(\Delta f, T_t)$; occasional uphill moves escape local minima and $T_t$ cools over time.
We track convergence, best FOM, and final design quality. Using OpenYield’s higher-fidelity circuit models with matched SPICE evaluation counts, comparisons are fair and reproducible while reflecting dominant failure mechanisms in manufactured arrays. With the best bank-level organization selected by FOM, we then proceed to bottom-level transistor tuning for SRAM bitcells.

\subsection{Bottom-Level: SRAM Bitcells}
\label{sec:sram_transistor}
As our transistor-level step, we size the standard 6T SRAM bitcell using OpenYield~\cite{openyield}. We adopt the same settings and algorithms as OpenYield: SPICE-based, PVT-aware Monte Carlo across process–voltage–temperature corners; worst-case aggregation of hold/read/write SNM, read/write delay, and dynamic power; and a robustness-centric figure of merit that increases with margin and penalizes power, delay, and area. Optimization follows OpenYield’s single- and multi-objective formulations and solver choices, ensuring fair, reproducible comparisons under matched evaluation budgets and delivering robust bitcell configurations aligned with the array organization selected at the architecture level.




\section{Experimental Evaluation}
\label{sec:results}

All experiments are performed on a workstation with an Intel Xeon Gold 6330 CPU (2.00\,GHz) and an NVIDIA A100 GPU (40\,GB). 
The proposed PEA-GNN is implemented in PyTorch and PyTorch Geometric (PyG) using the Adam optimizer. 
Datasets are automatically generated within the OpenACMv2 framework based on the Nangate45\,nm open-cell library, where each synthesized design is evaluated under a 5\,ns clock period and a 10\,fF output load using OpenROAD and OpenSTA. 

\subsection{Accuracy and Efficiency of PEA-GNN}
As shown in \autoref{tab:gnn_accuracy_comparison}, PEA-GNN closely matches EDA ground truth across all metrics. For 8-bit designs, all targets achieve MSEs below $10^{-3}$, with MRED/NMED errors of 2.1\%/1.8\% and Delay/Area/Power errors under 0.3\%. All outputs reach $R^2{>}0.94$. For 16-bit designs, accuracy remains high despite the larger space: MRED/NMED errors are 4.7\%/2.3\%, PPA deviations stay below 0.25\%, and $R^2{>}0.95$, confirming the scalability of the stage-wise and hierarchical encoding. PEA-GNN also provides significant speedups: 0.26,s vs.\ 37,s for 8-bit (142$\times$) and 0.25,s vs.\ 116,s for 16-bit (464$\times$), enabling rapid evaluation of large design spaces and efficient Pareto-front exploration.

Overall, PEA-GNN attains near-EDA accuracy while accelerating design-space evaluation by several orders of magnitude, serving as an effective surrogate for approximate-multiplier optimization in OpenACMv2.

\begin{table}[tbp]
\centering
\caption{Inference accuracy and runtime comparison of 8-bit and 16-bit PEA-GNN models.}
\label{tab:gnn_accuracy_comparison}
\setlength{\tabcolsep}{4pt}
\resizebox{\columnwidth}{!}{
\begin{tabular}{lcccccc}
\toprule
\multirow{3}{*}{\textbf{Metric}} 
& \multicolumn{3}{c}{\textbf{8-bit}} 
& \multicolumn{3}{c}{\textbf{16-bit}} \\
\cmidrule(lr){2-4} \cmidrule(lr){5-7}
& \textbf{MSE ($\times10^{-4}$)} & \textbf{MRE (\%)} & \textbf{R$^2$}
& \textbf{MSE ($\times10^{-4}$)} & \textbf{MRE (\%)} & \textbf{R$^2$} \\
\midrule
MRED  & 0.52 & 2.17 & 0.998 & 2.94 & 4.73 & 0.977 \\
NMED  & 9.25 & 1.84 & 0.996 & 2.66 & 2.30 & 0.959 \\
Delay & 6.52 & 0.03 & 0.969 & 7.39 & 0.22 & 0.917 \\
Area  & 1.77 & 0.11 & 0.991 & 3.65 & 0.12 & 0.978 \\
Power & 2.34 & 0.30 & 0.989 & 0.86 & 0.25 & 0.968 \\
\midrule
Runtime (s) 
& \multicolumn{3}{c}{0.26 (GNN) vs.\ 37 (EDA) (\textbf{142$\times$})}
& \multicolumn{3}{c}{0.25 (GNN) vs.\ 116 (EDA) (\textbf{464$\times$})} \\
\bottomrule
\multicolumn{7}{l}{\footnotesize EDA evaluation is performed by OpenROAD + OpenSTA + VCS.}
\end{tabular}}
\end{table}

\subsection{Approximate Multiplier}

\subsubsection{Top-Level Results.}

\begin{figure}[tbp]
    \centering 

    \begin{minipage}[b]{0.48\linewidth} 
        \centering
        \includegraphics[width=\linewidth]{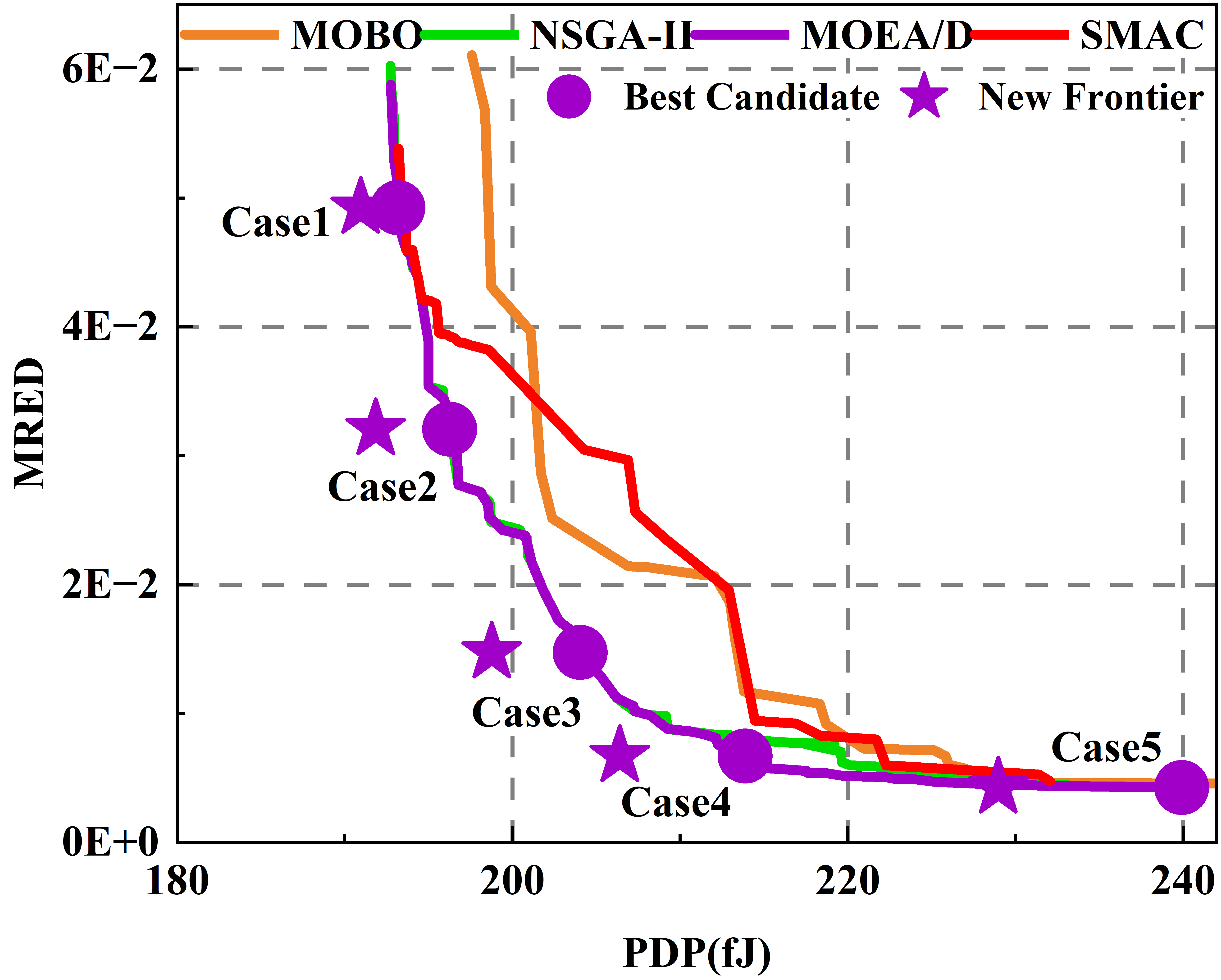}
        \captionof{figure}{Arch.-level Pareto Frontier of 8-bit multiplier obtained by optimizers.}
        \label{fig:level1_pareto_8}
    \end{minipage}
    \hfill 
\begin{minipage}[b]{0.48\linewidth}
    \centering
    \setlength{\tabcolsep}{3pt}
    \resizebox{\linewidth}{!}{
    \begin{tabular}{l c cc cc}
    \toprule
    \multirow{2}{*}{\textbf{Mult.}} &
    \multirow{2}{*}{\textbf{MRED}} &
    \multicolumn{2}{c}{\textbf{PDP(fJ)}} &
    \multicolumn{2}{c}{\textbf{PSNR}} \\
    \cmidrule(lr){3-4} \cmidrule(lr){5-6}
    & & \textbf{Lv-1} & \textbf{Lv-2} &
    \textbf{Test0} & \textbf{Test1} \\
    \midrule
    Base & 2.40E-03 & \multicolumn{2}{c}{484} & 69.81 & 70.59 \\
    Case1 & 5.88E-02 & 193 & 191 & 47.03 & 46.01 \\
    Case2 & 3.21E-02 & 196 & 192 & 49.08 & 48.96 \\
    Case3 & 1.47E-02 & 204 & 199 & 51.03 & 51.38 \\
    Case4 & 6.70E-03 & 214 & 206 & 54.94 & 56.79 \\
    Case5 & 4.25E-03 & 240 & 229 & 58.57 & 64.31 \\
    \bottomrule
    \multicolumn{6}{l}{\footnotesize Test0: Cameraman + Boat, Test1: Cameraman + Lake} \\
    \end{tabular}}
    \captionof{table}{Image PSNR comparison under accuracy constraints.}
    \label{tab:image_psnr}
\end{minipage}
\vspace{-3pt}
\end{figure}

\begin{figure}[tbp]
    \centering 

    \begin{minipage}[b]{0.50\linewidth} 
        \centering
        \includegraphics[width=\linewidth]{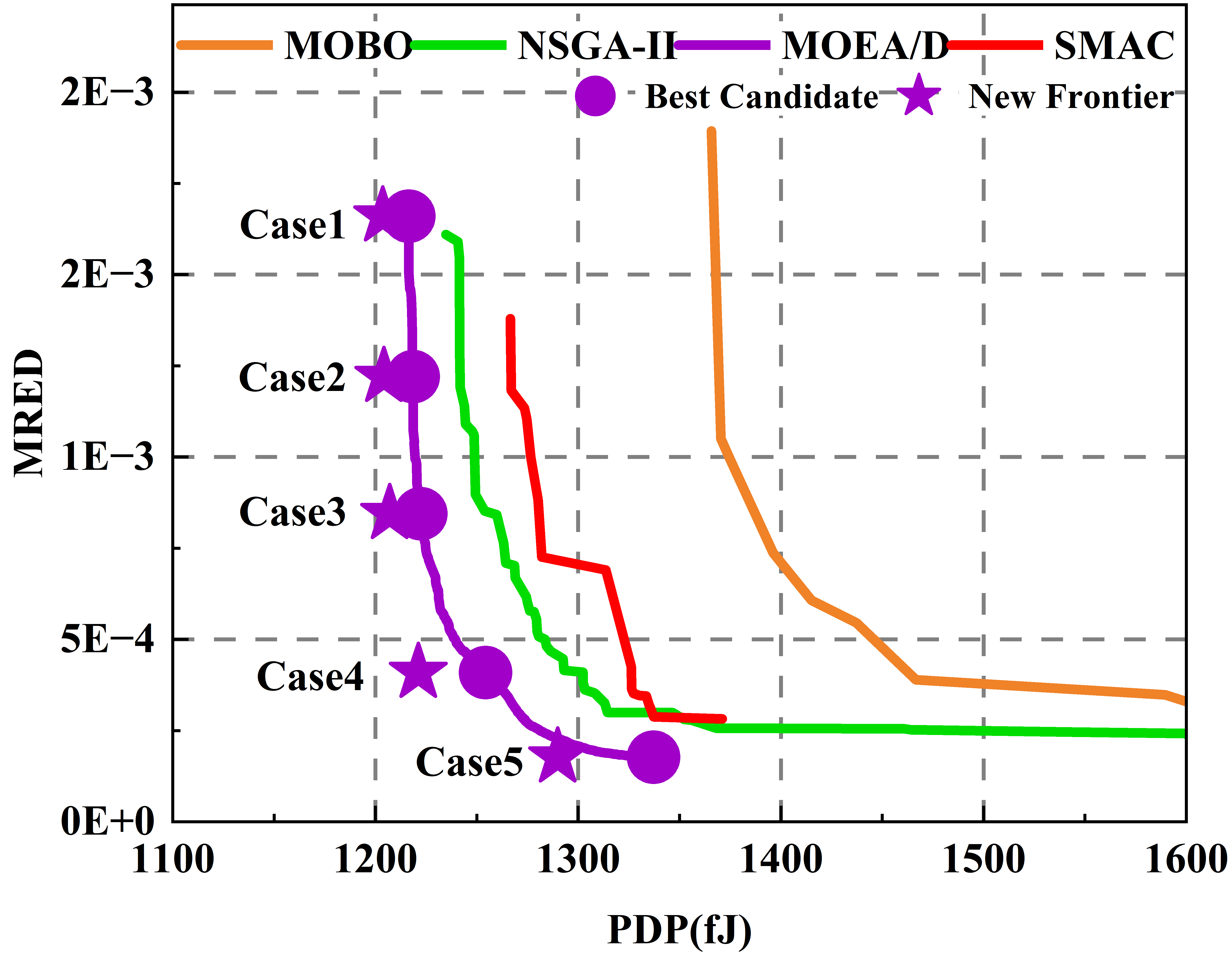}
        \captionof{figure}{Arch.-level Pareto frontiers of 16-bit multiplier obtained by optimizers.}
        \label{fig:level1_pareto_16}
    \end{minipage}
    \hfill 
    \begin{minipage}[b]{0.46\linewidth} 
        \centering
        \setlength{\tabcolsep}{2pt}
        \resizebox{\linewidth}{!}{
        \begin{tabular}{l c cc c c}
        \toprule
        \multirow{2}{*}{\textbf{Mult.}} &
        \multirow{2}{*}{\textbf{MRED}} &
        \multicolumn{2}{c}{\textbf{PDP(fJ)}} &
        \multirow{2}{*}{\textbf{Top-1}} &
        \multirow{2}{*}{\textbf{Top-5}} \\
        \cmidrule(lr){3-4}
        & & \textbf{Lv-1} & \textbf{Lv-2} & & \\
        \midrule
        Base  & 1.27E-10  & \multicolumn{2}{c}{3874} & 66.6 & 86.7 \\
        Case1     & 1.66E-03  & 1216 & 1203 & 65.7 & 86.3 \\
        Case2     & 1.22E-03  & 1219 & 1204 & 65.8 & 86.3 \\
        Case3     & 8.44E-04  & 1222 & 1207 & 65.1 & 86.2 \\
        Case4     & 4.09E-04  & 1254 & 1221 & 66.5 & 86.7 \\
        Case5     & 1.77E-04  & 1337 & 1289 & 65.7 & 86.4 \\
        \bottomrule
        &   &  &  &  & \\
        \end{tabular}}
        \captionof{table}{MRED, PDP, and Top-$k$ (\%) comparison under accuracy constraints.}
        \label{tab:mred_pdp_acc_twotier}
    \end{minipage}
\vspace{-3pt}
\end{figure}

\autoref{fig:level1_pareto_8} and \autoref{fig:level1_pareto_16} present the Pareto frontiers of the 8/16-bit multipliers under four optimizers. 
MOEA/D exhibits the most stable convergence; therefore, five representative points (Case1--Case5) are selected for further analysis. Recall ACCO: we treat application-level error budgets (here measured by MRED) as \emph{hard constraints} in Level-1. Each Case represents a distinct MRED budget; the architecture search selects designs that \emph{meet} the budget while minimizing PDP.

Tab. \ref{tab:image_psnr} summarizes their performance on an image blending task. 
Under ACCO, moving along the frontier from Case1 (relaxed budget, higher MRED) to Case5 (tight budget, lower MRED), PSNR improves accordingly, confirming that accuracy-constrained selection translates to task-level quality. 
The Base design (Yang1 compressor) achieves the highest PSNR but with an excessively low MRED and large energy overhead; ACCO avoids over-precision by enforcing the budget and selecting energy-optimal architectures. 
For each budget, Level-2 device tuning further reduces PDP while preserving PSNR (see Lv-1 vs. Lv-2 in Tab.~\ref{tab:image_psnr}), satisfying ACCO’s requirement that device-level changes do not violate the accuracy constraint.

For the 16-bit case, Tab. \ref{tab:mred_pdp_acc_twotier} reports CIFAR-10 inference under ACCO budgets spanning roughly an order of magnitude in MRED. Despite this spread, Top-1/Top-5 accuracy varies by less than 1\%, highlighting NN error tolerance and the value of budgeted approximation. Several Pareto-optimal points match the Base design’s accuracy while significantly reducing PDP, and Level-2 tuning consistently lowers PDP within each budget (Lv-1 vs. Lv-2), with accuracy unchanged.

\subsubsection{Bottom-Level Results.}

To further improve implementation efficiency, we apply transistor-level optimization to each compressor. 
\autoref{fig:level2_all}~(left) presents the results of four optimization algorithms on the Sabetz compressor, where MOEA/D achieves the best Pareto-front convergence and the most favorable PDP--area trade-off. 
Motivated by this observation, we adopt MOEA/D to optimize all eight approximate compressors individually, as shown in \autoref{fig:level2_all}~(right). 
In every case, the optimized Pareto front lies to the left of its corresponding standard-size baseline (diamond markers), indicating that transistor-level tuning delivers significant and consistent improvements in the PDP--area space.
Importantly, device-level tuning respects ACCO: architectural error (MRED) remains unchanged across Level-2, so PDP reductions do not compromise the accuracy budget.

\begin{figure}[tbp]
    \centering
    \includegraphics[width=\linewidth]{figs/level-2all.png}
    \caption{Transistor-level Pareto frontiers: across optimizers on the Sabetz compressor~\cite{sabetzadeh2019majority} (left); across compressors using MOEA/D (right). Rhomboids are reference points.}
    \label{fig:level2_all}
   \vspace{-3pt}
\end{figure}

\subsection{SRAM Macro}

\begin{figure}[t]
\centering
\begin{minipage}[t]{0.47\linewidth}
    \centering
    \includegraphics[width=\linewidth]{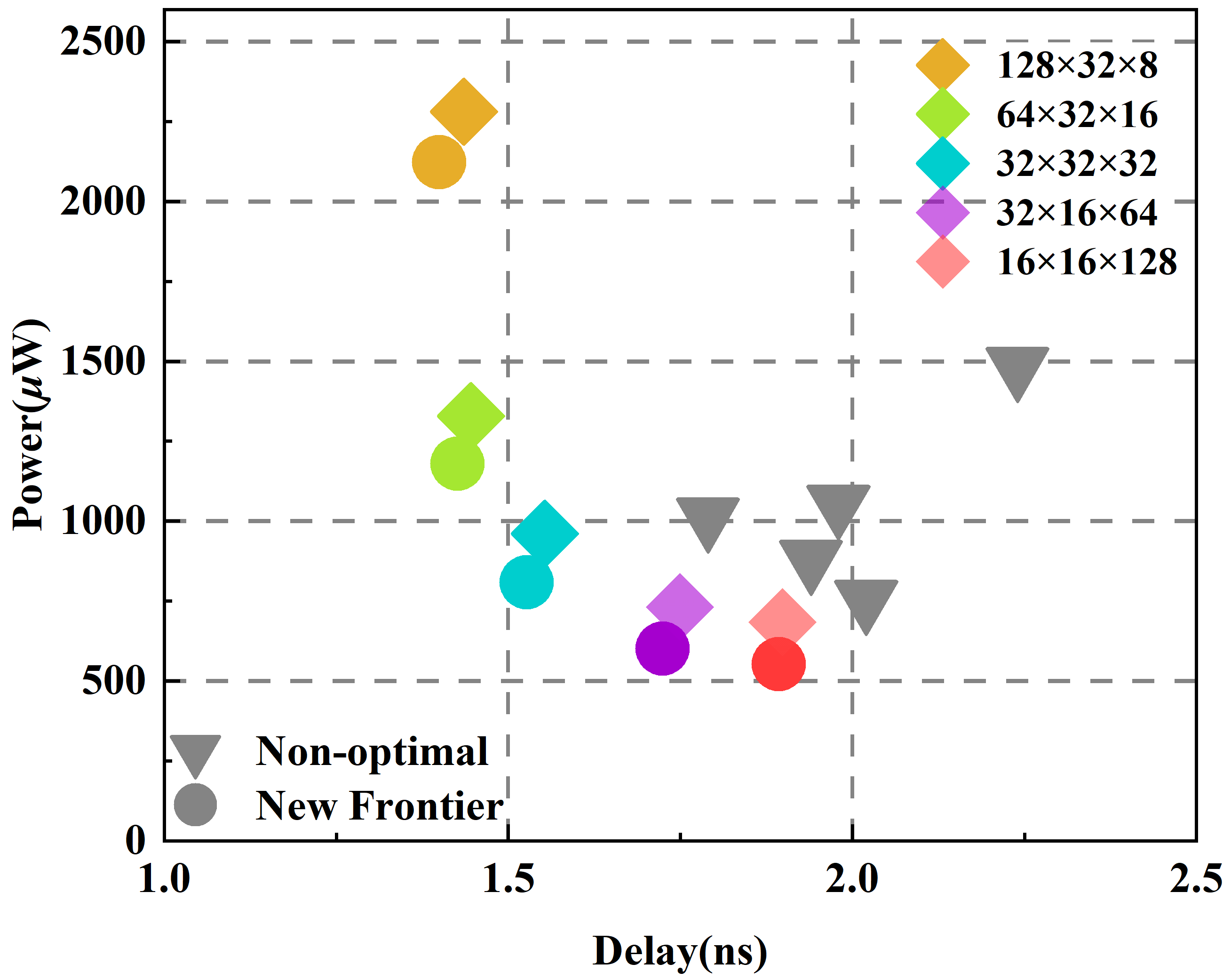}
    \caption{Arch.-level SRAM Pareto frontiers under capacity constraints.}
    \label{fig:level1_SRAM}
\end{minipage}
\hfill
\begin{minipage}[t]{0.48\linewidth}
    \centering
    \includegraphics[width=\linewidth]{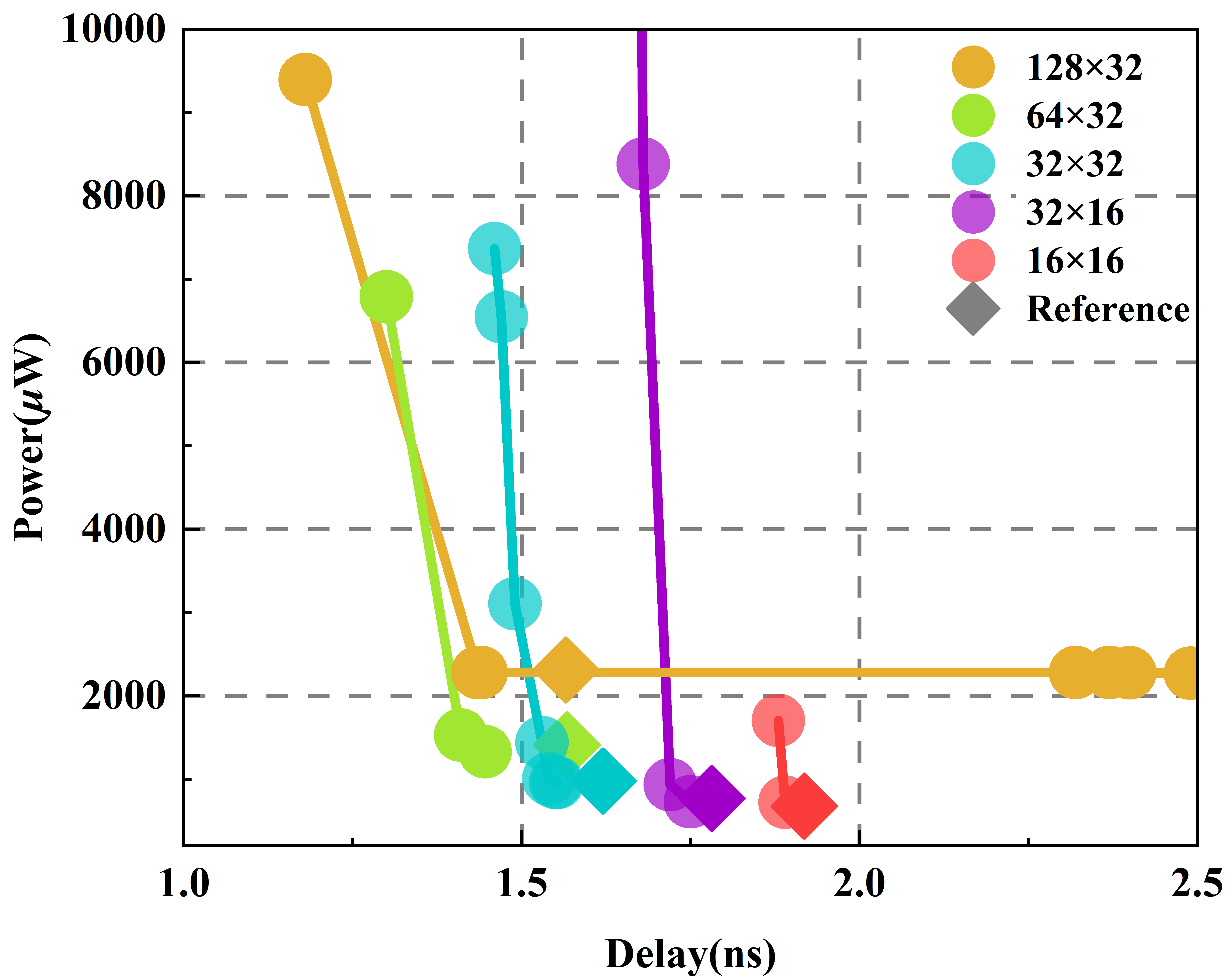}
    \caption{Trans.-level Pareto frontiers across the best bank configurations.}
    \label{fig:level2_SRAM}
\end{minipage}
\vspace{-3pt}
\end{figure}

As shown in \autoref{fig:level1_SRAM}, the architecture-level exploration scans bank organizations by varying row/column splits, array counts, and mux ratios under a fixed capacity budget. Smaller arrays (fewer rows/columns) exhibit lower energy but longer delay dominated by periphery, such as data multiplexing and pre-decoding; increasing row number reduces the array count, yielding lower delay but higher dynamic power due to larger load on bitlines. Across optimizers, the resulting power--delay trade-off traces a concave Pareto front, with the highest-FOM designs clustering near the ``elbow'' where periphery cost and bitline switching are balanced. These shortlisted banks are passed directly to Level-2 transistor sizing.

At the bottom level (\autoref{fig:level2_SRAM}), transistor sizing across the best bank configurations yields only limited reductions in terms of delay and dynamic power compared with the architecture-level gains. The non-ideal peripheral circuits restrict safe sizing windows, resulting in wider exploration failing in SPICE simulations during optimization. 
Both \autoref{fig:level1_SRAM} and \autoref{fig:level2_SRAM} demonstrate that only optimizing bitcell devices offers incremental headroom but does not materially shift the power--delay frontier. Consequently, architecture-level choices dominate the SRAM PPA under ACCO, while device-level tuning provides fine-grained adjustments.

\subsection{Putting All Together}

Under explicit MRED budgets, the two-level ACCO jointly improves energy efficiency without degrading task accuracy. Level-1 architecture exploration \emph{establishes} the accuracy--PDP frontier at each budget (\autoref{fig:level1_pareto_8},~\autoref{fig:level1_pareto_16}; Tab.~\ref{tab:image_psnr},~\ref{tab:mred_pdp_acc_twotier}). Level-2 transistor tuning then \emph{shifts} the frontier left—reducing PDP while keeping PSNR and Top-$k$ accuracy unchanged—so the combined flow delivers lower energy at equal or tighter error budgets across bit-widths and compressor families.
For SRAM macros specifically, bottom-level sizing delivers more area reductions—despite negative impact on delay and power—though detailed area results are omitted due to space limitations.

\section*{Conclusion \& Limitations}
\label{sec:conclusion}
OpenACMv2 advances ACCO for DCiM with a two‑level flow that shifts the accuracy–PDP Pareto front toward lower energy under fixed accuracy budgets. Level‑1 performs PEA‑GNN–guided architecture search (compressors, SRAM) for 8/16‑bit designs; Level‑2 applies variation/PVT‑aware sizing to shortlisted logic and memory, reducing PDP while preserving PSNR and Top‑$k$ accuracy. Together, the flow yields energy‑optimal designs at the required precision across compressor families and bit‑widths.

Limitations. (i) Surrogate fidelity and coverage across PVT corners; (ii) narrow objective scope focused on PDP and accuracy—omits throughput, leakage, IR drop, and routing congestion; (iii) incomplete backend signoff integration (place-and-route, parasitic extraction, crosstalk); (iv) limited depth in SRAM macro and bitcell/periphery co-optimization. 

{
\bibliographystyle{ACM-Reference-Format}
\bibliography{IEEEabrv, refs}
}

\end{document}